\documentclass[acmtog]{acmart}

\usepackage{multirow}
\usepackage{colortbl}

\definecolor{warningcolor}{HTML}{FFFAD4} %  - Warning
\definecolor{rankonecolor}{HTML}{FFDADA}   % - Best
\definecolor{ranktwocolor}{HTML}{CAEEFB}   %  - Second

\newcommand{\rankone}{\cellcolor{rankonecolor}}
\newcommand{\ranktwo}{\cellcolor{ranktwocolor}}

\copyrightyear{2026}
\acmYear{2026}
\setcopyright{cc}
\setcctype{by}
\acmConference[SA Conference Papers '26]{SIGGRAPH Asia 2026 Conference Papers}{December 01--04, 2026}{Kuala Lumpur, Malaysia}
\acmBooktitle{SIGGRAPH Asia 2026 Conference Papers (SA Conference Papers '26), December 01--04, 2026, Kuala Lumpur, Malaysia}
\acmDOI{10.1145/3829340.3842244}
\acmISBN{979-8-4007-2842-6/2026/12}

\begin{document}
% Title portion
\title{SiZeUp: Fast 3D Proxy from Aerial Images via Depth Ordinal Loss}

\author{Wenjun Zhou}
\orcid{0000-0003-1790-4201}
\affiliation{%
     \department{CSSE}
	\institution{Shenzhen University}
	\country{China}	
}
\email{wenjun.9707@gmail.com}
\author{Yunshan Li}
\affiliation{%
   % \department{VCC, CSSE}
	\institution{Shenzhen University}
	\country{China}	
}
\email{yunshanli2001@gmail.com}
\author{Qiaoyu Zhu}
\affiliation{%
    %\department{VCC, CSSE}
	\institution{Shenzhen University}
	\country{China}	
}
\email{qiaoyuzhu.stu@gmail.com}
\author{Weidan Xiong}
\affiliation{%
    %\department{VCC, CSSE}
	\institution{Shenzhen University}
	\country{China}	
}
\email{xiongweidan@gmail.com}

\author{Hao Zhang}
\affiliation{%
 \institution{Simon Fraser University}
 % \city{Prague}
 \country{Canada}}
\email{haoz@cs.sfu.ca}

\author{Daniel Cohen-Or}
% \affiliation{%
%  \institution{Tel Aviv University}
%  \department{School of Engineering}
%  \city{Charlottesville}
%  \state{VA}
%  \postcode{22903}
%  \country{Israel}
% }
\affiliation{%
 \institution{Shenzhen University}
 \country{China}}
\email{cohenor@gmail.com}

\author{Hui Huang}
%\email{hhzhiyan@gmail.com}
\authornote{Corresponding author: Hui Huang (hhzhiyan@gmail.com)}
\affiliation{%
     \department{Guangdong Provincial Key Laboratory of Visual Media and Multidimensional Intelligence, CSSE}
	\institution{Shenzhen University}
	\country{China}	
}

\begin{abstract}
We present \emph{SiZeUp}, a fast and scalable approach for constructing large-scale 3D urban proxy models directly from calibrated oblique aerial imagery. Our method adopts a height-from-footprint representation, reducing 3D building abstraction to a low-dimensional optimization problem in which building footprints are extruded by a single height parameter. To enable efficient and robust height estimation, we introduce an \emph{ordinal depth consistency loss} that enforces agreement between the relative depth ordering of rendered proxies and depth priors predicted by a monocular depth model. This is realized through a differentiable renderer that maps parametric building proxies into multi-view depth images, allowing gradients to be propagated from depth supervision to building heights. Our ordinal formulation produces stable optimization in practice and avoids explicit feature matching or dense point cloud reconstruction. Rather than relying on metric depth, which can be unreliable under monocular scale ambiguity, our ordinal depth consistency loss operates on relative depths, providing a more reliable signal across views. Combined with an efficient dynamic view selection, our approach achieves a 23-52$\times$ speedup over state-of-the-art proxy reconstruction pipelines while maintaining comparable proxy-level coverage and volume consistency, making it well suited for large-scale urban modeling tasks.
\end{abstract}





\begin{teaserfigure}
  \centering
  \includegraphics[width=\linewidth]{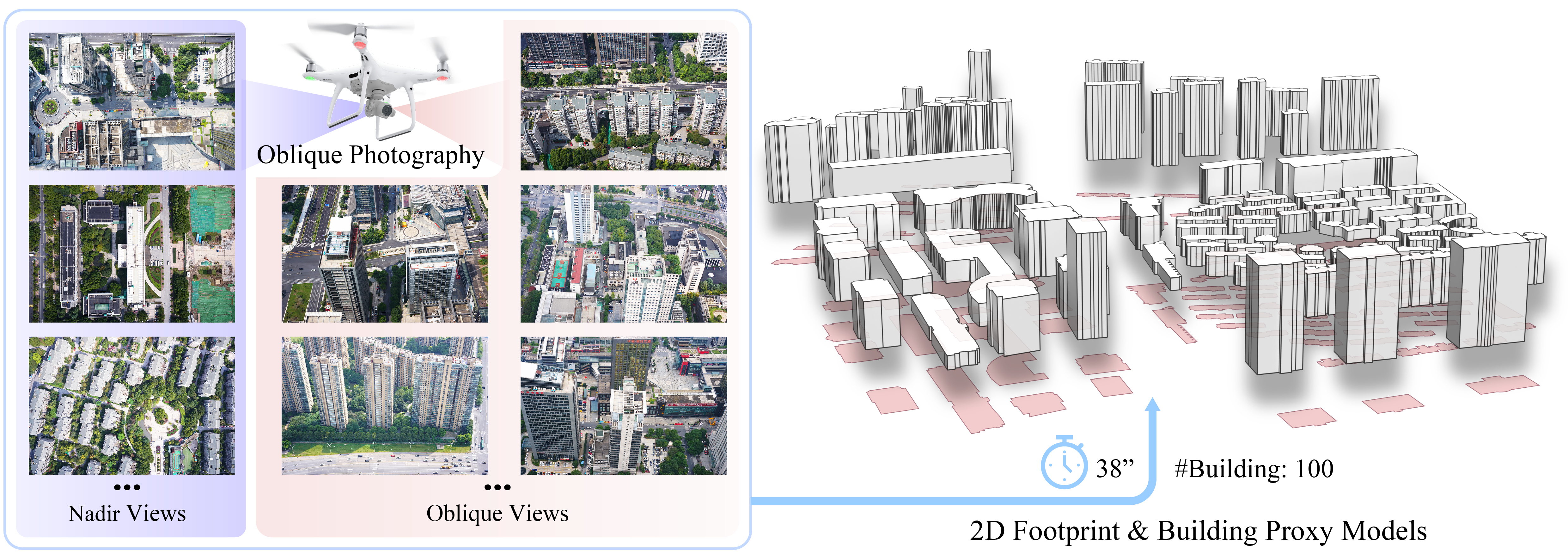}
  \vspace{-7mm}
  \caption{From oblique aerial images, SiZeUp extracts building footprints (shown in pink) from nadir views and lifts them into compact 3D proxies by estimating building heights. Efficiency of our method is owing to direct image-to-height differentiable renderer using only a small portion of the images via view selection.
  }
  \label{fig:teaser}
\end{teaserfigure}

\maketitle

\section{Introduction}

Automated reconstruction of 3D urban environments from aerial imagery remains a challenging problem in computer graphics, with a wide range of practical applications across planning, simulation, and analytics for real-estate, AR/VR, as well as digital twins and smart cities~\cite{kelly2017bigsur,Nan2017polyfit,ProxyRecon24}.
When dealing with large-scale urban scenes, a coarse-to-fine reconstruction strategy is typically employed where in the first stage, a \emph{3D proxy} is obtained to guide the subsequent fine-grained geometry recovery, e.g., for drone path planning \cite{Hepp2018Plan3D,DronePath21,roberts2017,liu_siga22,smith2018}.
Such a proxy model is a simplified yet structurally faithful and meaningful geometric representation to capture the spatial arrangement and core volumetric properties of buildings in the scene.
Besides serving as a reconstruction prior, the 3D proxies can directly benefit downstream tasks where structural correctness and scalability matter more than visual realism.

Naturally, proxy construction should be significantly faster than dense urban reconstruction.
However, conventional methods to proxy estimation often resort to point cloud extraction from images first~\cite{ArcPro25}, while many downstream applications that require proxies produce them as abstractions over dense models obtained by Structure-from-Motion (SfM)~\cite{snavely2006photo}, followed by multi-view stereo (MVS)~\cite{furukawa2009accurate}.
These indirect approaches incur a lengthy workflow with high computational costs and data requirements; their results are highly dependent on the quality of the reconstructed point clouds, which require high-overlap imagery to yield plausible intermediate 3D representations~\cite{mur2015orb,engel2017direct}.
Moreover, RTK-equipped UAV platforms are widely adopted in modern oblique aerial photogrammetry, and such systems commonly record or calibrate camera poses as part of the capture process.
Hence, calibrated images serve as a natural and practical starting point to directly optimize building proxies in image space, while bypassing the SfM/MVS reconstruction required by point-cloud-based proxy methods.

\begin{figure*}[ht] 
    \centering
    \includegraphics[width=\linewidth]{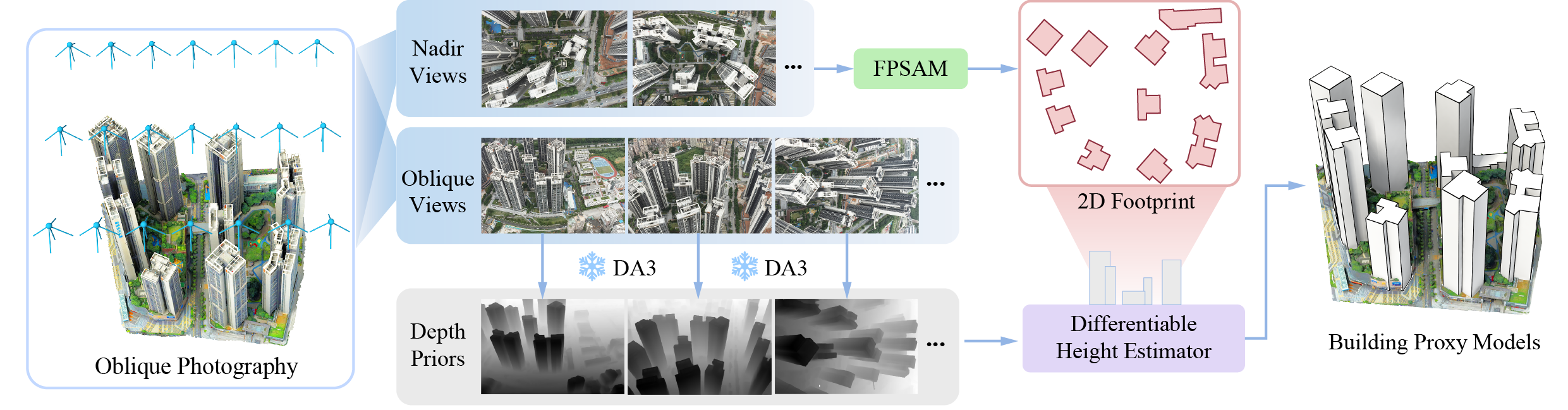}
    \vspace{-6mm}
    \caption{Overview of our urban proxy estimation method, called \textit{SiZeUp}. Given calibrated aerial images, our \textit{SiZeUp} first extracts 2D footprints from nadir views, then rapidly lifts the 2D silhouettes based on a small set of dynamically selected oblique views.
    }
    \vspace{-4mm}
    \label{fig:overview}
\end{figure*}

In this paper, we introduce a method to extract 3D proxies \emph{directly} from calibrated multi-view oblique aerial images with \emph{high efficiency}. 
Our proxies are modeled by extruding 2D building footprints into vertical volumes, reducing each building geometry estimation to a single height inference, and as such, our 3D proxy solution is coined \emph{SiZeUp}.
Our key observation is that such a reduced representation turns efficient proxy reconstruction from a general 3D recovery problem into a constrained image-space alignment problem, where \emph{relative} depth structure provides a more reliable signal than metric depth values, enabling an image-space loss that can well tolerate the inevitable imprecisions from depth or 3D estimates over a small set of input images. This motivates the development of a differentiable renderer that can be guided by an \emph{ordinal depth objective} that aligns rendered proxy geometry with noisy depth priors.

Specifically, the differentiable renderer maps building heights into multi-view depth spaces, and exploits the ordinal structure inherent in the depth priors to define the gradient through a novel \emph{ordinal consistency loss}. 
This loss enforces that the relative depth ordering between two pixels in the rendered depth matches the ordering given by a prior depth map.
For pixel pairs with sufficient prior depth contrast, it penalizes cases where the rendered depth reverses or weakly respects this ordering using a hinge loss with a margin, encouraging a minimum separation.
Averaging this penalty over sampled pixel pairs yields a view-level loss that preserves correct depth ordering rather than absolute depth values.

To complete our proxy solution pipeline, we first extract the 2D footprints based on a learned footprint extractor, which combines semantic and appearance cues from SAM~3~\cite{carion2025sam} with geometric cues from Depth Anything~3~(DA3)~\cite{lin2025depth}.
This is followed by building height estimation via the differentiable renderer, where we resort again to DA3 to provide depth supervision.
The multi-view depth ordering, together with parallax and occlusion cues, yields meaningful gradients without requiring any high-cost dense surface reconstruction.
Overall, the high efficiency of our framework can be attributed to not only our breaking down of the proxy estimation into searches over low-dimensional parameter spaces, i.e., 2D footprints plus height extrusion, but also a fast \emph{dynamic view selection} algorithm which leads to light-weight optimization over only few hundred input images.

We show quantitative and qualitative results from SiZeUp and evaluate each of its components: footprint extraction, view selection, and building height estimation. 
Importantly, even though our method produces coarser 3D proxies compared to low-poly meshes~\cite{ProxyRecon24}, convex hulls~\cite{Open3D}, and recent program-based architectural abstractions such as ArcPro~\cite{ArcPro25}, it achieves the best efficiency-quality tradeoff among all baselines. Specifically, SiZeUp attains a 23-52$\times$ speedup over these baselines on real scenes while maintaining comparable proxy-level coverage and volume consistency, though at the cost of higher point-level surface error due to its coarser LOD1 representation.

\section{Related Works}
\label{sec:related}

Since existing literature on 3D reconstruction is vast, we will mainly cover related works on \emph{structured} scene extraction in this section, especially those involving urban proxy models.

\paragraph{3D proxy construction}
Most existing methods construct proxies based on precomputed 3D data, such as dense reconstructed meshes, from which geometric abstractions can be obtained via mesh simplification~\cite{garland1997surface} or piecewise-planar approximation~\cite{cohen2004variational, verdie2015lod, salinas2015structure, kelly2017bigsur}.
When working with point clouds, existing approaches detect geometric primitives and assemble them into compact polyhedral building proxies~\cite{Monszpart2015rapter, Nan2017polyfit, fang2020connect, bauchet2020kinetic, bouzas2020structure, Pan2022Efficient, Guo2022, ProxyRecon24, ArcPro25, citygo}.
Inverse procedural modeling can recover procedural descriptions from raw 3D data~\cite{mathias2011procedural, toshev2010detecting}, typically relying on robust plane extraction~\cite{raguram2012usac} and other carefully designed architectural priors. However, requiring high-fidelity 3D meshes or point clouds as input can be prohibitively expensive and impractical to obtain at city scale.

In the absence of such data, architectural proxies are typically created through manual modeling or interactive procedural tools~\cite{sinha2008interactive}.
Image-based reconstructions~\cite{ceylan2012factored} exploit line structures and repetitive facade elements to infer coarse urban buildings, while procedural grammars enable users to synthesize large collections of proxies~\cite{parish2001procedural, schwarz2015advanced}.
Nevertheless, grammar authoring remains a specialized task requiring domain expertise, and these methods either demand strong facade regularity assumptions or substantial manual intervention for complex scenes.

In this paper, we focus on reconstructing large-scale urban scene proxies directly from oblique RGB imagery, bypassing the need for 3D inputs.
In practice, urban building proxies are often generated through simple footprint extrusion~\cite{DronePath21,li20243d}, where precise 2D footprint segmentation and height estimation from images constitute the core technical challenges.

\paragraph{Footprint extraction}
Traditional image-based methods primarily rely on geometric cues such as edges~\cite{canny1986computational}, shadows~\cite{huertas1988detecting, zhou2020offsite}, and vanishing points~\cite{caprile1990using}, typically assuming near-nadir viewing angles.
For large-scale scenes, indirect methods project 3D meshes or point clouds obtained by SfM/MVS onto the ground plane to derive the urban footprint. 
In the absence of such 3D data, a common strategy couples semantic instance segmentation~\cite{liu2019building, wagner2020u} with auxiliary remote-sensing products, for instance radar-guided fusion~\cite{Wang2024Sentinel} or orthorectified satellite mosaics, to approximate footprint regions.
In parallel, monocular and satellite imagery have motivated learning-based extractors that emit footprints or vectorized outlines directly from pixels, including graph-based polygon predictors~\cite{PolyWorld}, sequence decoders for building polygons~\cite{Pix2Poly2025}, and roof--footprint-coupled reconstruction under multi-level supervisions~\cite{li20243d} and prompt-driven offset-building models~\cite{OBM}.
Constrained by the lack of footprint-aware segmentation models, these methods are generally limited in dealing with specialized sensing modalities.
In contrast, our input is simply photographic images captured with oblique photography, exhibiting significant variations in building orientations and severe building occlusions. 
Existing methods struggle to handle such complex imagery due to their reliance on idealized assumptions or limited generalization capabilities.
To fill this gap, we introduce a novel footprint extractor that integrates semantics, depth, and texture cues. Moreover, we construct a dedicated near-nadir aerial footprint benchmark for training and validation.

\paragraph{Height and depth estimation}
Building height can be inferred by feature matching and image cues.
\citet{zhou2020offsite} generates 3D proxies by exploiting viewpoint and illumination constraints, requiring satellite imagery with pronounced shadows as input.
\citet{li20243d} reconstructs 3D buildings from monocular remote-sensing images via multi-level supervision.
While effective in controlled settings, these methods rely on task-specific annotations or sensing modalities (e.g., satellite imagery with consistent lighting), limiting their generalization to oblique aerial photography, where repetitive urban textures and varying perspectives introduce ambiguity.

An alternative strategy estimates the building height from depth or nadir images~\cite{Wang2024Sentinel}. Recent advances in generic depth and 3D foundation models have significantly improved accuracy.
Monocular depth objectives have long sought to reduce their sensitivity to ambiguous global scale and shift. Eigen et al.~\cite{eigen2014depth} introduced scale-invariant regression in log-depth space, Chen et al.~\cite{chen2016single} learned single-image depth from pairwise ordinal annotations using a ranking loss, and Ranftl et al.~\cite{ranftl2020towards} developed robust scale- and shift-invariant objectives for training on heterogeneous depth datasets. Unlike these methods, which learn depth predictors from metric or relative-depth supervision, we keep the pretrained monocular depth estimator fixed and transfer only its within-view ordering to building-height parameters defined in the metric proxy coordinate system through calibrated multi-view differentiable rendering.
Notably, Depth Anything~3~\cite{lin2025depth} achieves state-of-the-art performance in monocular depth estimation.
However, these models are trained on curated datasets that differ from oblique aerial inputs, and their results still suffer from scale ambiguity~\cite{arampatzakis2023monocular}.
Consequently, their outputs are better suited as geometric priors rather than as precise height constraints.

Generally speaking, existing methods do not fully exploit the rigid geometric properties inherent to building footprints. Most approaches treat footprints as post-processing masks rather than fundamental anchors for reconstruction itself.
To address these challenges, we propose an ordinal-based height estimation method that organically integrates footprint constraints with depth priors. This approach mitigates scale ambiguity while leveraging the inherent geometric relationships between footprints and building heights, enabling robust estimation in oblique aerial imagery.

\begin{figure*}[ht] 
    \centering
    \includegraphics[width=\linewidth]{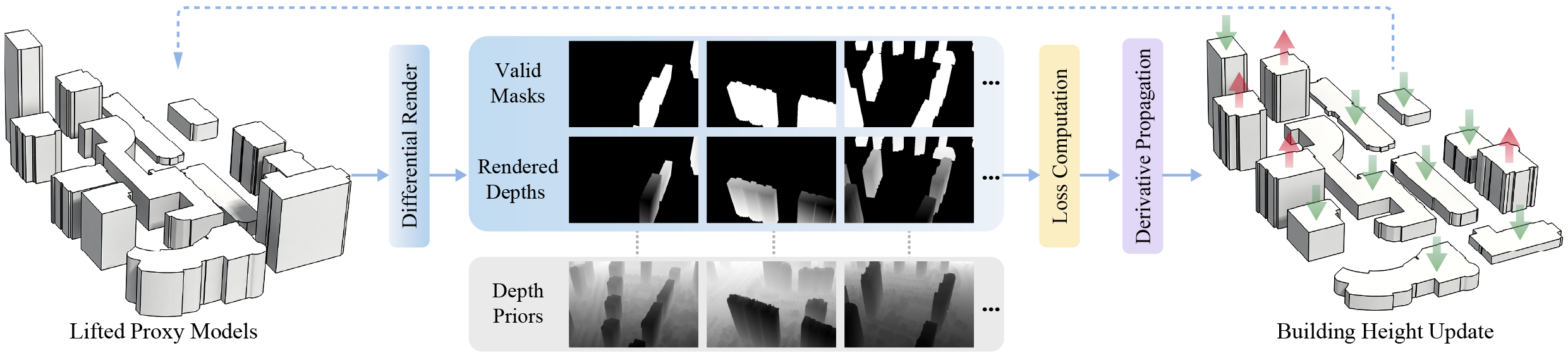}
    \vspace{-6mm}
    \caption{Overview of our differentiable building height estimation method.}
    \vspace{-4mm}
    \label{fig:height-estimation}
\end{figure*}

\section{Method}
\label{sec:method}

We take as input calibrated oblique aerial images $\{\mathcal{I}_k\}$ captured by an RTK-equipped UAV, with their camera poses recorded onboard during acquisition.
Our goal is to directly estimate building proxies $\{\mathcal{P}_i\}$ by extruding building footprints to specific heights $\{h_i\}$.
Fig.~\ref{fig:overview} presents the overview of our method.
For an arbitrary scene, we classify a view as nadir if the angle between its optical axis and the downward vertical is below $15^\circ$; all remaining views are treated as oblique.
Then, we apply a segmentation model, \textit{FPSAM} to the nadir views to obtain building footprint masks.
These masks are then vectorized into polygonal footprints $\{\mathcal{F}_i\}$.
Next, depth priors from oblique views are extracted using DA3.
Finally, building heights $\{h_i\}$ are estimated by minimizing an objective function that measures the discrepancy between the rendered proxy depths and the corresponding depth priors.
Below, we provide a more detailed description of each step.
\subsection{Footprint Extraction}
\label{sec:footprint_extraction}

Existing footprint extraction methods struggle with low-altitude oblique aerial imagery. Near-orthographic methods \cite{PolyWorld,Pix2Poly2025} rely on the premise that visible contours represent footprints, yet this fails under perspective distortion, facade visibility, and self-occlusion. While offset-based methods \cite{li20243d,OBM} resolve geometric displacement, they introduce a significant annotation bottleneck by requiring structured supervision of roofs, footprints, or offsets.

To address these limitations, we introduce \textit{FPSAM}, a footprint extractor that fuses texture, semantic, and depth-aware geometric cues derived from two frozen foundation models. 
The SAM~3 branch~\cite{carion2025sam} provides two pixel-level features from the near-nadir image $\mathcal{I}_k$: a texture feature from its image encoder and FPN, and a semantic feature from its text-guided grounding branch with the prompt ``building''. 
Meanwhile, the DA3 branch~\cite{lin2025depth} extracts a depth-aware feature from its ViT-L backbone, providing coarse geometric structure. 
These feature maps are projected using lightweight $1\times1$ convolutions, concatenated, and then decoded by a trainable U-Net and mask decoder, where only the projectors and decoder are updated under footprint-mask supervision. 
By combining the localization cues from SAM~3 with the geometric information from DA3, \textit{FPSAM} achieves improved discrimination of ground-contact footprints from roof boundaries, facade silhouettes, and visible building masks (see supp. for more details).

To train and evaluate our footprint extractor, we introduce \textit{VertiFP}, a near-nadir aerial benchmark annotated with ground-contact footprints. 
The training set comprises $7$ real urban scenes, approximately $10^3$ buildings, $4\times10^3$ images, and $3\times10^4$ footprint instances, while the test set contains $2$ scenes, roughly $10^2$ buildings, $4\times10^2$ images, and $2\times10^3$ instances. 
Each footprint instance corresponds to one annotated building per image. 
Given a near-nadir image, \textit{FPSAM} predicts footprint masks, which are then vectorized into closed polygonal footprints $\{\mathcal{F}_i\}$~\cite{douglas1973algorithms} and remain fixed for the subsequent height optimization.

\subsection{Building Height Estimation}

Given the building footprints $\{\mathcal{F}_i\}$, we construct building proxies by extruding each footprint polygon vertically according to its building height.
The values of building heights $\{h_i\}$ are initialized randomly from a normal distribution within a scene-dependent range, $[h_\mathrm{min},h_\mathrm{max}]$. The range is determined by the scene elevation and the UAV acquisition altitude.
%In this stage, we aim at optimizing the heights $\{h_i\}$ for all the buildings in the scene. 
In this stage, a differentiable optimization method is proposed to update the heights $\{h_i\}$ for all the buildings in the scene, constrained by the input aerial views. The pipeline of our method is shown in Fig.~\ref{fig:height-estimation}.

\begin{figure}[t]
    \centering
    \includegraphics[width=\linewidth]{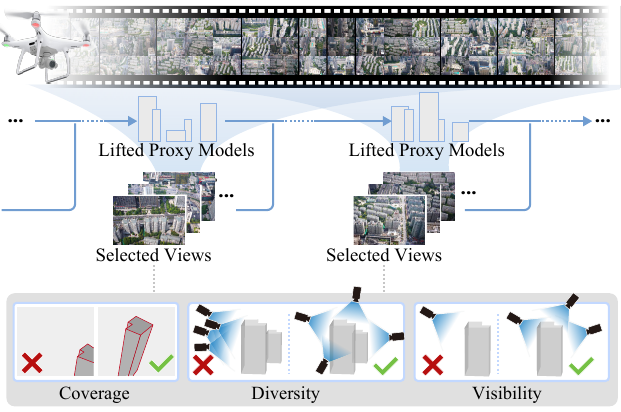}
    \vspace{-6mm}
    \caption{Overview of our dynamic view selection strategy. Given all available oblique views, we periodically select a fixed number of informative views to supervise the proxy update.}
    \label{fig:dvs}
    \vspace{-4mm}
\end{figure}

\paragraph{Dynamic View Selection}
Oblique aerial photogrammetry is typically planned with dense overlap to ensure feature-based reconstruction of dense representations.
However, the majority of the captured photos remain redundant or weakly informative for proxy reconstruction. 
To enable efficient height optimization given scene proxies, we begin with a \emph{dynamic view selection} (DVS) strategy to select a small subset of views $\{\mathcal{I}\}_o^\ast$ from the oblique views $\{\mathcal{I}\}_o$.
The selected views should:
(i) provide high coverage of building surface,
(ii) exhibit spatial diversity and variation in viewing angles~\cite{zhou2020offsite}, 
and (iii) ensure that each building is observed by at least $k$ views.
A heuristic view selection method is introduced, guided by the above three criteria.  
More specifically, we first assess the quality of the view $\mathcal{I}_k$ in $\{\mathcal{I}\}_o$ with two metrics after projecting $\mathcal{I}_k$ onto current proxies: the number of visible buildings $Q_N$ and the total area of covered building surface $Q_A$. 
Views with quality values smaller than percentile-based thresholds $\tau_N$ or $\tau_A$ will be discarded. We use $\tau_N = \tau_A = 50\%$ in the experiments.
Next, we select views from the remaining images using farthest point sampling~\cite{FPS}, ensuring complete ground coverage and spatial diversity.
To ensure sufficient multi-view constraints, we further enforce a minimum exposure requirement, iteratively adding views that improve building-wise coverage until each building is observed from $k$ viewpoints (we let $k=3$).
As the proxy heights evolve, we re-evaluate and update the selected view subset  $\{\mathcal{I}\}_o^\ast$ every $3$ epochs based on the updated scene, as shown in Fig.~\ref{fig:dvs}.

In practice, DVS reduces the number of views involved in height optimization, improving the efficiency of the overall pipeline while retaining informative multi-view constraints.
In our real-scene experiments, only 10\%--27\% of the available views are used throughout optimization (see Table~\ref{tab:real_scene_eval}).
A more detailed validation of DVS is provided in the supplemental material.

\paragraph{Depth priors}
A depth map is capable of providing geometric cues in estimating building heights.
We make use of a monocular depth estimator, Depth Anything~3~\cite{lin2025depth}, to compute depth priors $\{\mathcal{D}_k\}$ of the views in $\{\mathcal{I}\}_o^\ast$.
Given current proxies at iteration $t$, following the camera parameters of each selected view $\mathcal{I}_k$, we render a depth map $\mathcal{R}_k$ and generate a binary mask $\mathcal{M}_k$ indicating all the visible building regions. We define $\Omega_k$ as the subset of pixels with non-zero value in $\mathcal{M}_k$. 
To make the depth prior and rendered depth maps comparable, we first normalize $\mathcal{D}_k$ via a median normalization~\cite{benchmark3d} over the valid building regions following:
\begin{equation}
\mathcal{D}_k' =
\frac{\mathcal{D}_k - \mathrm{median}(\mathcal{D}_k, \Omega_k)}
     {\mathrm{MAD}(\mathcal{D}_k, \Omega_k) + \varepsilon},
\end{equation}
\begin{equation}
    \mathrm{MAD}(\mathcal{D}_k, \Omega_k) =
    \mathrm{median}
    \bigl( \lvert
    \mathcal{D}_k - \mathrm{median}(\mathcal{D}_k , \Omega_k)
    \rvert , \Omega_k  \bigr),
\end{equation}
where $\mathrm{median}(A, \Omega)$ selects the subset pixels $\Omega$ from matrix $A$ and returns their median value, and $\varepsilon$ is a relatively small value.
Similarly, we normalize the rendered depth map $\mathcal{R}_k$ to obtain $\mathcal{R}_k'$.

\begin{figure}[t]
   \centering
   \includegraphics[width=\columnwidth]{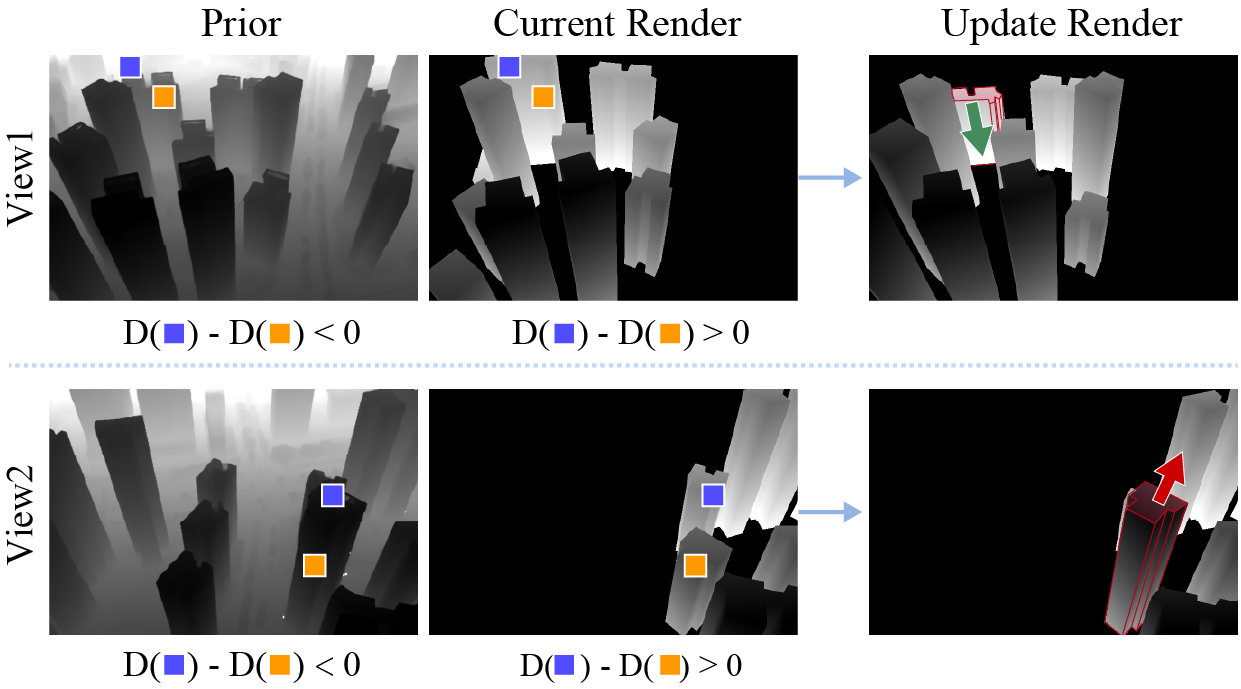}
   \vspace{-4mm}
   \caption{Visualization of ordinal consistency. Left: depth prior and rendered depth exhibit mismatched ordering. Right: such inconsistencies drive height updates toward ordinal agreement with the depth prior.} 
   \label{fig:ordinal_consistency}
   \vspace{-4mm}
\end{figure}

\paragraph{Objective function}
Recovering accurate metric depth from monocular depth priors remains challenging in practice because their metric scale is inherently ambiguous. Unlike scale-invariant depth regression~\cite{eigen2014depth,ranftl2020towards} or ordinal supervision used to train a depth predictor~\cite{chen2016single}, we keep Depth Anything~3 fixed and transfer only the pairwise ordering of its predictions to the rendered proxy depth.
We therefore encourage the depth ordering in $\mathcal{R}_k$ to match that in $\mathcal{D}_k$, rather than directly comparing their depth values.
As illustrated in Fig.~\ref{fig:ordinal_consistency}, the colored markers denote sampled pixel pairs whose ordinal relationships are inconsistent between the depth prior and the rendered depth.
Such inconsistencies can arise not only between buildings, but also between a building and its surrounding context when an overestimated proxy height causes the rendered valid mask to cover background pixels.
Penalizing violations will decrease the building height in View 1, and increase the building height in View 2. Such an ordinal can be computed using the difference among pixel values in a depth map.
For each selected view $\mathcal{I}_k$, we define the normalized depth differences of a pixel pair $(p,q)$ from $\Omega_k$ as: 
\begin{equation}
\Delta_p^k(p,q) = \mathcal{D}_k'(q) - \mathcal{D}_k'(p),
\end{equation}
\begin{equation}
\Delta_r^k(p,q) = \mathcal{R}_k'(q) - \mathcal{R}_k'(p).
\end{equation}

For every view used in an optimization iteration, we randomly resample $2048$ pixel pairs $(p,q)$ from its valid region $\Omega_k$.
Then, to avoid degenerate constraints, we discard candidate pairs $(p,q)$ with negligible difference: $|\Delta_p^k(p,q)| < \tau_d$ (we set $\tau_d = 0.05$). 
The remaining pairs $(p,q)$ are recorded in $P_k$.

To regularize the solution and avoid trivial matches, we enforce a minimum separation between ordered pixels.
A surrogate loss function, $l^k(p,q)$, which resembles the hinge loss~\cite{hingeloss}, is employed to measure the ordinal \emph{inconsistencies} among pixel pairs $(p,q)$, while also penalizing weakly ordered pairs that satisfy the ordering but fall within too small of a margin:
\begin{equation}
l^k(p,q) =
\max\{0,\; m - \mathrm{sign}\bigl(\Delta_p^k(p,q)\bigr) \cdot \Delta_r^k(p,q) \},
\end{equation}
where $m$ is the margin, and setting $m = 0.5$ achieves stable performance empirically. The depth difference provides a discrete ordinal label: $\operatorname{sign}(\Delta_d^k) \in \{-1,0,1\}$, while $\Delta_r^k$ varies smoothly with the proxy heights through the differentiable renderer.

We compute the ordinal violations of a view $\mathcal{I}_k$ as the average of $l^k(p,q)$ over all the pixel pairs in $P_k$: 
\begin{equation}
L^k =
\frac{1}{|P_k|}
\sum_{(p,q) \in P_k}
l^k(p,q),
\end{equation}
where $|P_k|$ is the number of pixel pairs in $P_k$.
The heights ${h_i}$ are shared across all selected views in $\{\mathcal{I}\}_o^\ast$. We aggregate the per-view ordinal violation into the final objective function:
\begin{equation}
L = \sum_{\mathcal{I}_k \in \{\mathcal{I}\}_o^\ast } L^k .
\end{equation}

Since $\mathrm{sign}(\Delta_p^k)$ is fixed and $\Delta_r^k$ depends differentiably on the heights via $\mathcal{R}_k'$, $L$ is piecewise differentiable with respect to $\{h_i\}$ and can be minimized efficiently with a gradient-based optimizer.
This multi-view coupling improves the conditioning of the optimization and avoids degenerate per-view solutions.

\begin{figure}[t]
    \centering
    \includegraphics[width=\linewidth]{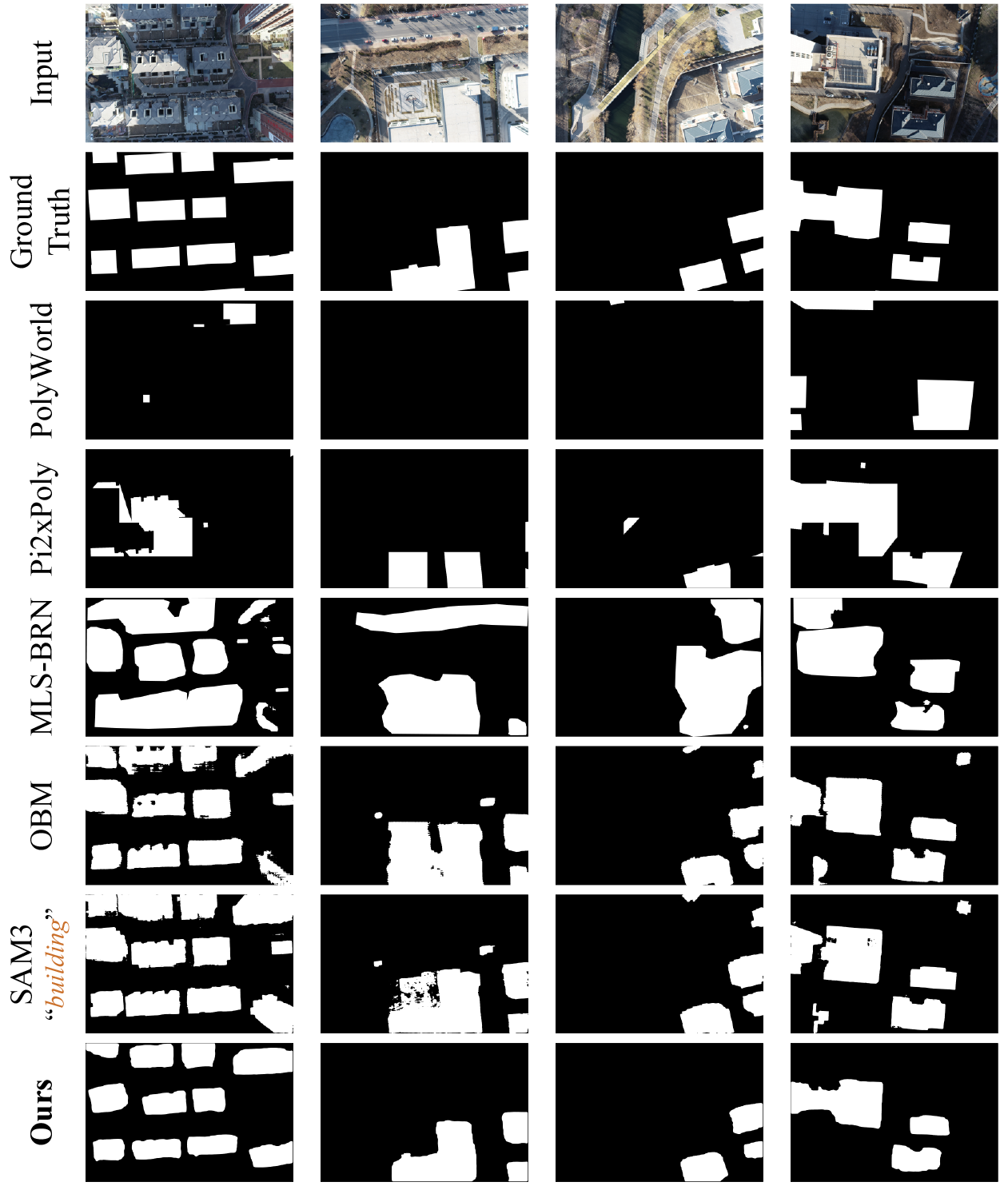}
    \vspace{-4mm}
    \caption{Qualitative footprint comparison on samples from the test split. From top to bottom: input images, ground-truth footprints, and predictions from PolyWorld, Pix2Poly, MLS--BRN, OBM, SAM~3, and our FPSAM.
    }
    \vspace{-8mm}
    \label{fig:fpsam}
\end{figure}

\begin{figure*}[t]
    \centering
    \includegraphics[width=\linewidth]{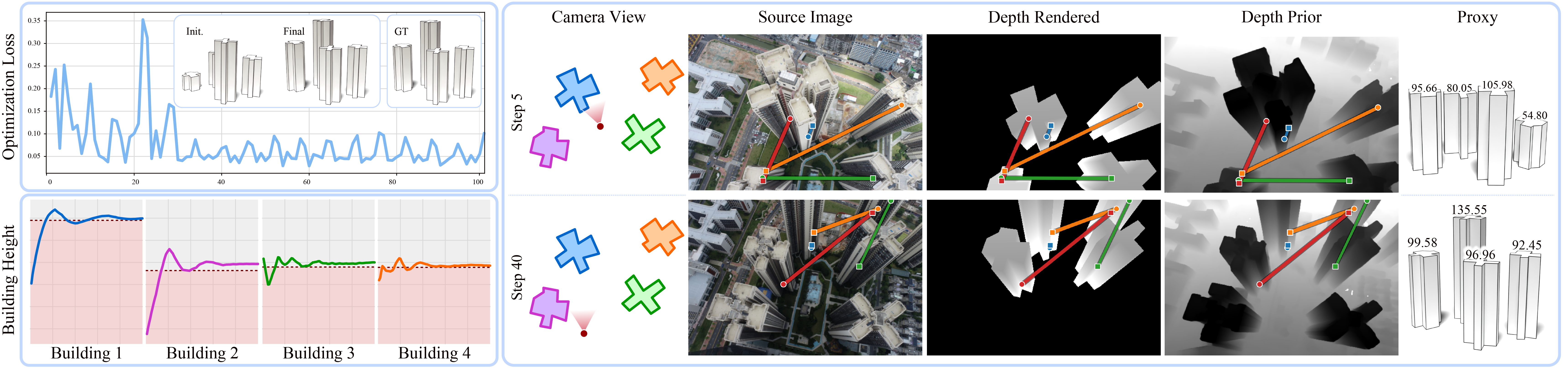}
    \vspace{-4mm}
    \caption{Detailed optimization process on the second mini-scene. Left: ordinal loss and per-building height trajectories. Right: selected optimization steps showing the camera layout, source image, rendered depth, depth prior, and current proxy with representative active ordinal pairs.}
    \vspace{-4mm}
    \label{fig:mini_scene_opt_vis_small}
\end{figure*}

\section{Experimental Results}

We evaluate the quality of footprint extraction, dynamic view selection, and height-based proxy reconstruction.
For footprint extraction, we follow the instance-level matching protocol of~\citet{chen2025sam} and report IoU, Precision, Recall, and F1-Score. 
For height estimation, we report absolute height error and relative height accuracy with respect to the reference building height, summarized by averages and high-percentile statistics.
For proxy-level comparison with point-cloud-based methods, we evaluate distances on uniformly sampled surface points from the reconstructed proxy and the ground-truth geometry.
\textit{Err.} measures nearest-neighbor distance from reconstructed samples to reference samples, while \textit{Comp.} measures reverse-direction coverage, i.e., the percentage of reference samples whose nearest reconstructed sample lies within a given threshold.
We report \textit{Comp.} 0.5m and \textit{Comp.} 1m, together with volumetric IoU.
Lower \textit{Err.} indicates better surface-level fitting, while higher \textit{Comp.} and IoU indicate better coverage and volume consistency.
Finer implementation details, baselines, and full metric definitions are provided in the supplemental material.

\begin{table}[t]
    \centering
    \caption{Quantitative comparison of building footprint extraction of aerial images on the test data. SAM~3 is prompted with \textit{"building"}.}
    % \vspace{-3mm}
   \label{tab:footprints_eval}
    \resizebox{\linewidth}{!}{
        \begin{tabular}{lcccc}
            \noalign{\hrule height 0.8pt}
            Method &
            IoU (\%)$\uparrow$ &
            Precision (\%)$\uparrow$ &
            Recall (\%)$\uparrow$ &
            F1-Score (\%)$\uparrow$  \\
            \midrule
            PolyWorld & 3.79  & 20.73  & 4.73  & 6.44 \\
            Pix2Poly  & 16.57  & 39.26  & 23.78  & 25.60 \\
            MLS-BRN  & 35.57  & 46.52  & 60.04  & 49.91 \\
            OBM  & 53.36  & 60.32  & 80.56  & 67.45 \\
            SAM3  & 58.98  & 62.34  & \textbf{89.86}  & 71.69 \\
            Ours  & \textbf{70.89}  & \textbf{81.49}  & 85.38  & \textbf{81.90} \\
            \noalign{\hrule height 0.8pt}
        \end{tabular}
    }
    \vspace{-6mm}
\end{table}

\subsection{Footprint Extraction}

We evaluate \textit{FPSAM} on the test set of \textit{VertiFP}, and compare it against representative footprint extraction and segmentation baselines, including the PolyWorld~\cite{PolyWorld}, Pix2Poly~\cite{Pix2Poly2025}, MLS--BRN~\cite{li20243d}, OBM~\cite{OBM}, and SAM~3~\cite{carion2025sam}(prompted with \textit{``building''}).
The qualitative comparison are visualized in Fig.~~\ref{fig:fpsam}, and the quantitative results are provided in Table~\ref{tab:footprints_eval}.

The results show that the performance of near-orthographic footprint methods degrades under oblique aerial imagery, i.e., PolyWorld and Pix2Poly. The visible contours no longer align with ground-contact footprints.
The MLS-BRN and OBM improve over these baselines by modeling roof-to-footprint offsets. 
Since the visible building mask often contains the footprint as a subset, the SAM~3 achieves high recall. However, SAM~3 shows lower precision, IoU, and qualitative results, due to its inaccurate footprint boundaries.
Meanwhile, our \textit{FPSAM} produces much cleaner and more compact masks, achieving the best IoU, precision, and F1-score.

\subsection{Height Optimization}
\label{sec:mini-ana}

Although the ordinal objective does not infer metric scale from monocular depth, height optimization remains metric: calibrated camera positions and ground-plane footprints are expressed in a common metric world coordinate system, in which the height variables and bounds $[h_\mathrm{min},h_\mathrm{max}]$ are defined.
The multi-view ordinal constraints therefore select metric heights within this calibrated geometry without regressing monocular metric depth (see Supplementary Sec.~9 for a comparison with direct pixel-wise $L_1$/$L_2$ regression).
We first analyze this optimization behavior on controlled mini-scenes (see the supplemental material for construction details).

Fig.~\ref{fig:mini_scene_opt_vis_small} shows a representative example.
The left panels track the ordinal loss and per-building height trajectories, with insets comparing the initial, final, and reference proxies.
We do not use the loss value to assess convergence. Instead, we monitor the average relative height change over all buildings between consecutive updates and regard the optimization as stable once this value falls below $0.5\%$.
This criterion is used only for stability assessment, since all experiments run for a fixed budget of $20$ epochs (see the supp.).
Although the loss fluctuates due to view-dependent visibility and imperfect depth priors, the heights move away from random initialization and settle into a stable LOD1 proxy.
The right panels show two optimization snapshots with representative active ordinal pairs overlaid on the source image, rendered depth, and depth prior. 
These pairs identify visible regions where the current proxy ordering disagrees with the prior ordering. 
Their nonzero losses accumulate gradients on the corresponding building heights, illustrating how SiZeUp updates heights from relative depth conflicts across views rather than fitting monocular metric depth values.

\begin{table}[t]
    \centering
    \caption{Statistics of height estimation on seven real-world scenes. \texttt{\#Sel.} denotes the number of views selected at each DVS step. \texttt{\#Used} denotes the total number of distinct selected views throughout the optimization. \texttt{Ratio} denotes the number of images used during optimization relative to the total number of available images.}
    \vspace{-2mm}
    \label{tab:real_scene_eval}
    \setlength{\tabcolsep}{5pt}
    \renewcommand{\arraystretch}{1.15} 
    \resizebox{\linewidth}{!}{
        \begin{tabular}{lcccccccc}
            \toprule 
            \multirow{2}{*}{Scene} & 
            \multicolumn{4}{c}{Optimization}&
            \multicolumn{2}{c}{Error (m) $\downarrow$} &
            \multicolumn{2}{c}{Accuracy (\%) $\uparrow$} \\   
            \cmidrule(lr){2-5} \cmidrule(lr){6-7}  \cmidrule(lr){8-9}           
            &  \#Sel. & \#Used & Ratio &Time (s) & Average & 90\%  & Average & 90\%  \\           
            \cmidrule(lr){1-1}  \cmidrule(lr){2-5} \cmidrule(lr){6-7}  \cmidrule(lr){8-9}          
            Res-1 & 50 & 101 & 0.27 & 19 & 3.28 & 7.19 & 93.45 & 88.34 \\
            Res-2 & 100 & 287& 0.17 & 38 & 7.47 & 15.93 & 85.95 & 61.07 \\
            Res-3 & 50 & 175 & 0.10 & 42 & 2.08 & 4.63 & 93.11 & 81.29  \\
            Res-4 & 100 & 298 & 0.12 & 49 & 4.27 & 7.49 & 89.67 & 77.19   \\
            Ind-1 & 50 & 147 & 0.10 & 29 & 1.28 & 2.56 & 92.47 & 66.22  \\
            Ind-2 & 100 & 261 & 0.21 & 33 & 6.48 & 14.35 & 77.85 & 44.09  \\
            City-1 & 256 & 873 & 0.17 & 118 & 4.84 & 6.52 & 85.19 & 65.30 \\           
            \bottomrule 
        \end{tabular}
    }
    \vspace{-4mm}
\end{table}

\begin{table*}[th]
\centering
\caption{
Quantitative comparison and runtime statistics for the 3D proxy construction of four real-world scenes. Runtime is divided into CCFront preprocessing and proxy construction/optimization. For each evaluation metric, the top-2 results are highlighted as \colorbox{rankonecolor}{best} and \colorbox{ranktwocolor}{second}, respectively.}

\label{tab:real_comp}

% =========================================================
% Adjustable table parameters
% =========================================================
% \setlength{\tabcolsep}{8.0pt}           % Horizontal column padding
\setlength{\tabcolsep}{0pt}
\def\BasePad{10pt}     % Scene, #Build. and Method
\def\TimePad{4pt}     % CCFront, Proxy/Opt. and Total
\def\MetricPad{12pt}   % Err., Comp. and IoU
\renewcommand{\arraystretch}{1.0}      % Vertical row spacing

\def\HeaderVShift{-0.8ex}
% =========================================================

\begin{tabular}{
    @{}
    >{\hspace{\BasePad}}c<{\hspace{\BasePad}}     % Scene
    >{\hspace{\BasePad}}c<{\hspace{\BasePad}}     % #Build.
    >{\hspace{\BasePad}}l<{\hspace{\BasePad}}     % Method
    >{\hspace{\TimePad}}c<{\hspace{\TimePad}}     % CCFront
    >{\hspace{\TimePad}}c<{\hspace{\TimePad}}     % Proxy/Opt.
    >{\hspace{\TimePad}}c<{\hspace{\TimePad}}     % Total
    >{\hspace{\MetricPad}}c<{\hspace{\MetricPad}} % Err. 90%
    >{\hspace{\MetricPad}}c<{\hspace{\MetricPad}} % Err. 95%
    >{\hspace{\MetricPad}}c<{\hspace{\MetricPad}} % Comp. 0.5m
    >{\hspace{\MetricPad}}c<{\hspace{\MetricPad}} % Comp. 1m
    >{\hspace{\MetricPad}}c<{\hspace{\MetricPad}} % IoU
    @{}
}

\noalign{\hrule height 0.8pt}
% Negative value moves the multirow headers downward

\multirow[c]{2}{*}[\HeaderVShift]{Scene}
& \multirow[c]{2}{*}[\HeaderVShift]{\#Build.}
& \multicolumn{1}{c}{
    \multirow[c]{2}{*}[\HeaderVShift]{Method}
}
& \multicolumn{3}{c}{Time (s)}
& \multicolumn{2}{c}{Err. (m) $\downarrow$}
& \multicolumn{2}{c}{Comp. (\%) $\uparrow$}
& \multirow[c]{2}{*}[\HeaderVShift]{IoU (\%) $\uparrow$} \\

\cmidrule(lr){4-6}
\cmidrule(lr){7-8}
\cmidrule(lr){9-10}

& & & CCFront & Proxy & Total
& 90\% & 95\%
& 0.5m & 1m
& \\

\midrule

% Res-1
\multirow{4}{*}{Res-1}
& \multirow{4}{*}{26}
& ConvexHull
& \multirow{3}{*}{442}
& 1
& 443
& 5.16
& 6.40
& 21.88
& 39.18
& 24.73 \\

& & ProxyRecon
& & 5
& 447
& \ranktwo 3.77
& \rankone 4.79
& \ranktwo 30.50
& \ranktwo 49.70
& \ranktwo 32.39 \\

& & ArcPro
& & 15
& 457
& 4.25
& 5.50
& 20.18
& 44.23
& 29.77 \\

\noalign{\vskip-\aboverulesep}
\cmidrule(lr){4-4}
\noalign{\vskip-\belowrulesep}

& & \textbf{SiZeUp}
& \textbf{--}
& \textbf{19}
& \textbf{19} ($23\times$)
& \rankone 3.73
& \ranktwo 5.38
& \rankone 57.97
& \rankone 76.77
& \rankone 49.18 \\

\midrule

% Res-2
\multirow{4}{*}{Res-2}
& \multirow{4}{*}{100}
& ConvexHull
& \multirow{3}{*}{1830}
& 1
& 1831
& 5.82
& 6.44
& 23.70
& 44.47
& 33.62 \\

& & ProxyRecon
& & 8
& 1838
& 4.78
& 6.51
& \ranktwo 33.63
& \ranktwo 58.78
& 36.26 \\

& & ArcPro
& & 134
& 1964
& \rankone 2.66
& \rankone 3.69
& 30.92
& 57.16
& \ranktwo 39.69 \\

\noalign{\vskip-\aboverulesep}
\cmidrule(lr){4-4}
\noalign{\vskip-\belowrulesep}

& & \textbf{SiZeUp}
& \textbf{--}
& \textbf{38}
& \textbf{38} ($48\times$)
& \ranktwo 4.52
& \ranktwo 6.07
& \rankone 41.21
& \rankone 66.65
& \rankone 42.87 \\

\midrule

% Ind-1
\multirow{4}{*}{Ind-1}
& \multirow{4}{*}{49}
& ConvexHull
& \multirow{3}{*}{1508}
& 1
& 1509
& 3.63
& 4.86
& 34.37
& 49.55
& 37.99 \\

& & ProxyRecon
& & 4
& 1512
& \ranktwo 2.51
& \ranktwo 2.97
& 37.63
& 51.92
& 42.74 \\

& & ArcPro
& & 42
& 1550
& \rankone 1.69
& \rankone 2.23
& \rankone 42.57
& \ranktwo 57.41
& \rankone 51.87 \\

\noalign{\vskip-\aboverulesep}
\cmidrule(lr){4-4}
\noalign{\vskip-\belowrulesep}

& & \textbf{SiZeUp}
& \textbf{--}
& \textbf{29}
& \textbf{29} ($52\times$)
& 3.33
& 4.70
& \ranktwo 37.86
& \rankone 64.68
& \ranktwo 43.12 \\

\midrule

% Ind-2
\multirow{4}{*}{Ind-2}
& \multirow{4}{*}{103}
& ConvexHull
& \multirow{3}{*}{1540}
& 1
& 1541
& \ranktwo 6.24
& \ranktwo 7.79
& 35.70
& 49.67
& 29.38 \\

& & ProxyRecon
& & 7
& 1547
& 13.46
& 16.26
& 26.14
& 43.53
& 29.54 \\

& & ArcPro
& & 117
& 1657
& \rankone 4.69
& \rankone 6.08
& \ranktwo 37.49
& \ranktwo 59.98
& \rankone 40.23 \\

\noalign{\vskip-\aboverulesep}
\cmidrule(lr){4-4}
\noalign{\vskip-\belowrulesep}

& & \textbf{SiZeUp}
& \textbf{--}
& \textbf{33}
& \textbf{33} ($47\times$)
& 8.46
& 10.40
& \rankone 48.34
& \rankone 61.56
& \ranktwo 35.36 \\

\noalign{\hrule height 0.8pt}
\end{tabular}

\end{table*}

\subsection{Evaluation on Real Scenes}

Next, we evaluate our method on four in-the-wild oblique photogrammetry scenes.
For each scene, we estimate building heights directly from raw images using our ordinal optimization pipeline and report the height estimation accuracy using the metrics defined above.
As shown in Table~\ref{tab:real_scene_eval}, SiZeUp produces accurate height estimates across a variety of urban layouts.

We compare SiZeUp against building proxy reconstruction methods, including ConvexHull~\cite{Open3D}, ArcPro~\cite{ArcPro25}, and ProxyRecon~\cite{ProxyRecon24}.
These methods operate on point clouds reconstructed from SfM (constrained by the same RTK metadata used directly by SiZeUp) that have been segmented into building instances.
Meanwhile, our novel SiZeUp directly generates proxies from images, bypassing point cloud reconstruction. 
To isolate reconstruction performance from instance segmentation, we use the footprints extracted by \textit{FPSAM} to obtain building instances for all point-cloud-based methods. 
As shown in Table~\ref{tab:real_comp}, SiZeUp exhibits higher point-level error (Err.) due to its extruded LOD1 representation, which omits fine roof/facade details. 
However, Comp. and IoU better reflect the proxy-level objectives of coverage and volume consistency. On these metrics, SiZeUp remains competitive with point-cloud-based methods, ranking among the top two in all real scenes while requiring only tens of seconds.

For the point-cloud baselines, we separate runtime into the SfM front end (CCFront) and the subsequent proxy reconstruction stage.
Our time covers depth estimation, dynamic view selection, and height estimation.
Our method requires only tens of seconds, corresponding to a 23-52$\times$ speedup (see Table~\ref{tab:real_comp}).
Ind-2 is the most challenging scene for surface-level accuracy, where both SiZeUp and the point-cloud baselines exhibit higher Err. Its higher flight altitude ($\sim$300m above the local mean ground elevation, versus $\sim$200m for Res-1/Res-2 and $\sim$100m for Ind-1) and the lowest image density and images-per-building ratio yield weaker geometric evidence. Nevertheless, SiZeUp remains top-2 in Comp. and IoU.

\begin{figure*}
	\centering
    \includegraphics[width=0.95\linewidth]{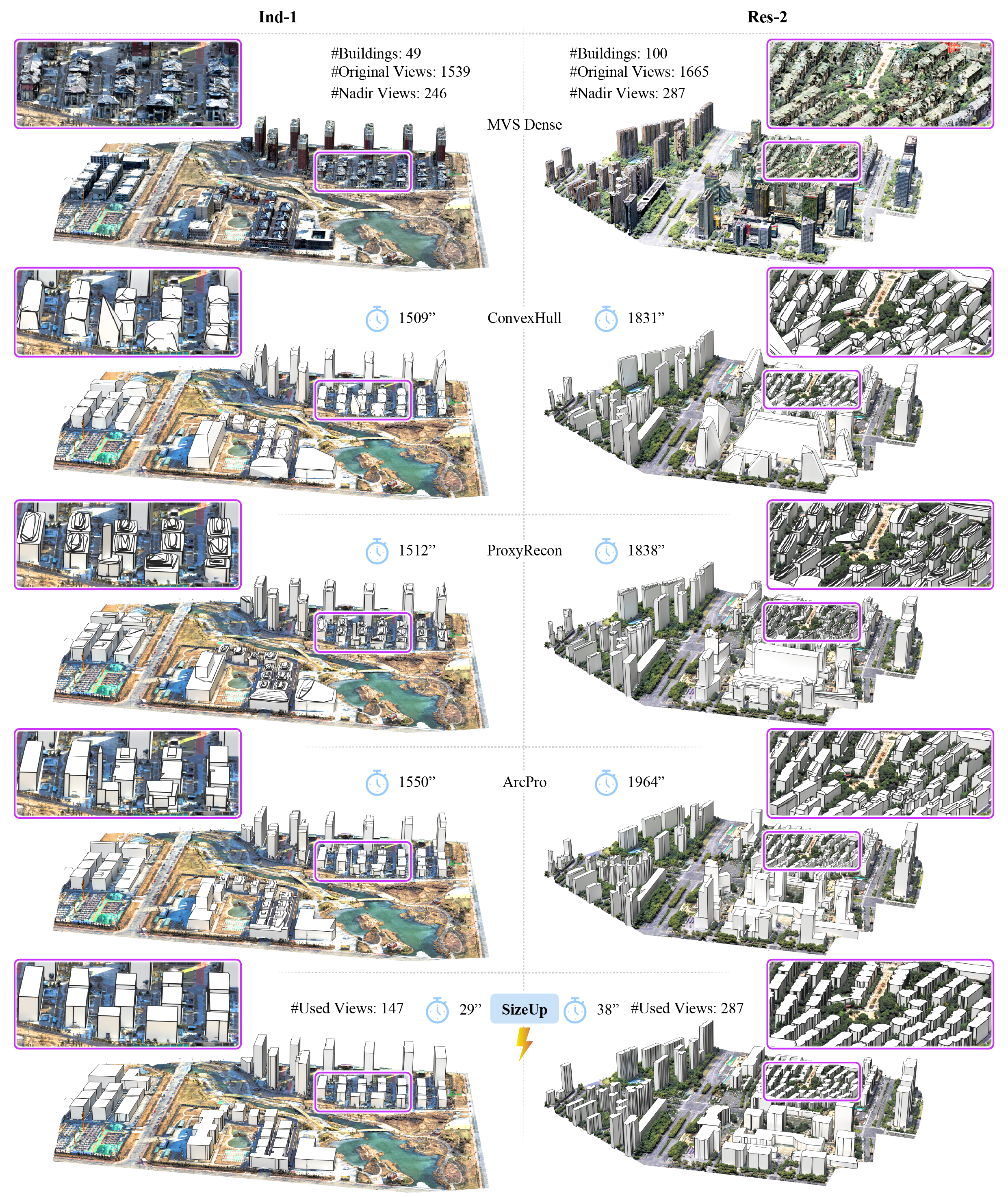}
    % \vspace{-2mm}
 	\caption{Comparison of constructed building proxies against MVS, ConvexHull, ProxyRecon, ArcPro, and our SiZeUp on two real scenes. The detailed comparisons are shown in the zoomed-in insets.
    Conventional methods operate on per-building point clouds obtained from SfM and instance segmentation, and reconstruct proxies through point-based geometric processing.
    Our SiZeUp directly infers building proxies from a small portion of images, yielding substantially lower runtime while preserving coherent  proxy-level coverage and volume consistency.}
	\label{fig:comparison}
\end{figure*}

Qualitative comparisons in Fig.~\ref{fig:comparison} further illustrate that SiZeUp maintains coherent scene-level structure without the need for SfM.
Together with the quantitative results, this demonstrates that direct image-to-proxy inference is a practical and viable alternative to conventional point-cloud-based proxy pipelines for urban modeling.
Additional visualizations are provided in the supplemental material.

\vspace{-4mm}
\subsection{Alternative Capture Pattern}

Standard UAV oblique-photogrammetry captures are often collected with grid-like flight routes, which naturally provide near-nadir views for footprint extraction in the full SiZeUp pipeline. 
Once building footprints are available, however, height optimization does not depend on this specific acquisition layout. 
It only requires views that provide informative depth-ordering cues, such as building visibility, occlusion relationships, and relative depth structure between buildings and their surrounding context; DVS then selects such views for ordinal supervision.
To test this property, we use a scene from UrbanScene3D~\cite{UrbanScene3D} with provided footprints, and generate non-grid camera views using the path planning strategy from~\cite{zhou2020offsite}. 
These views follow task-specific trajectories with varying viewpoints, distances, and flight heights, rather than a uniform overhead grid. 
Using the same optimization pipeline and settings as in the main real-scene experiments, SiZeUp reconstructs coherent LOD1 proxies under this capture pattern, as shown in Fig.~\ref{fig:campus_vis}. 
This suggests that the ordinal height optimization is driven by multi-view relative depth constraints rather than implications specific to standard oblique-photogrammetry flight route patterns.

\section{Conclusion and Discussion}

We introduced \textsc{SiZeUp}, a framework for efficiently producing urban 3D proxy models from oblique aerial imagery by estimating building heights over extracted footprints.
More broadly, this work is grounded in the view that many urban modeling tasks do not require full geometric reconstruction, but can be addressed more effectively by identifying an appropriate proxy representation and designing tools tailored to that reduced formulation. 
By deliberately constraining the representation and optimizing only the dominant structural degrees of freedom, we show that a focused, task-specific formulation can achieve substantial efficiency gains while preserving coherent proxy-level coverage and volume consistency.
This perspective highlights the importance of aligning representation, supervision, and optimization with the actual requirements of the task, rather than defaulting to increasingly complex and computationally expensive models.

\begin{figure}[ht]
   \centering
   \includegraphics[width=\columnwidth]{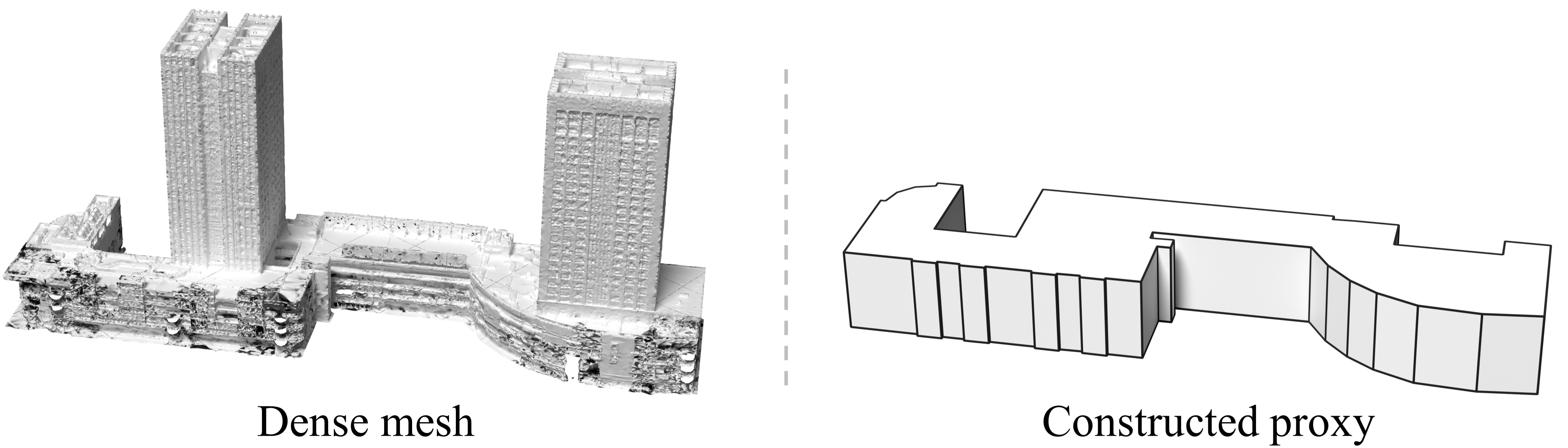}
    \vspace{-6mm}
    \caption{Building with multi-layer or stepped structures produced by footprint extrusion yields a less accurate proxy.}
    \vspace{-2mm}
   \label{fig:limitations}
\end{figure}

\begin{figure*}
	\centering
    \includegraphics[width=0.98\linewidth]{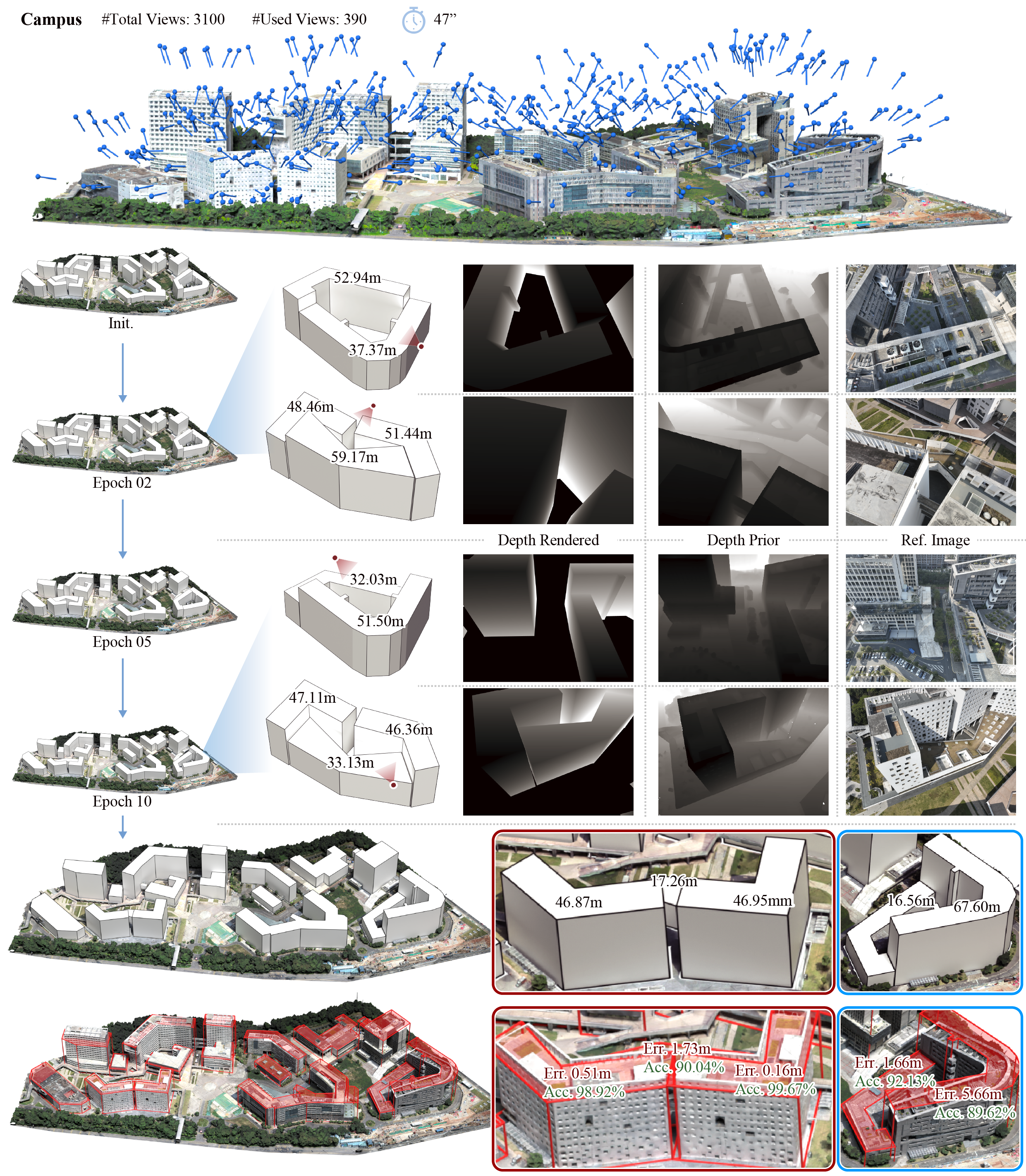}
    % \vspace{-6mm}
	\caption{
    Height optimization under a non-grid flight route and capture pattern. 
    Top: selected calibrated views and the ground-truth 3D model. 
    Middle (Init.~ to Epoch 10): optimization process of two representative local regions, showing rendered depths, depth priors, reference images, and the corresponding proxy heights at different stages.  
    Bottom: final constructed proxies and their overlay on the 3D model. 
    Zoomed regions show representative building-level results with optimized heights, absolute height errors (Err.), and relative accuracies (Acc.).
    }
	\label{fig:campus_vis}
\end{figure*}

Our approach represents buildings as vertical extrusions of planar footprints. 
When a complex building is modeled by a single footprint, SiZeUp cannot fully capture stepped roofs, multi-layer structures, or non-vertical details, leading to less accurate proxies (see Fig.~\ref{fig:limitations}).
This limitation can be partially mitigated by decomposing complex buildings into finer footprint components with separate heights, as illustrated in Fig.~\ref{fig:campus_vis}.
The quality of the results further depends on footprint accuracy and the availability of reliable ordinal depth cues.
Future work could explore richer proxy representations while preserving a low-dimensional formulation, as well as joint optimization of footprints and heights. Incorporating additional weak geometric cues may further stabilize scale, and extending this task-driven philosophy to other urban elements could enable broader proxy-level scene modeling.

% \section*{Acknowledgments}
\begin{acks}
This work was supported in part by National Key R\&D Program of China (2024YFB3908500, 2024YFB3908502), ICFCRT (W2441020), Guangdong Basic and Applied Basic Research Foundation (2023B151\allowbreak5120026), Shenzhen Science and Technology Program (KJZD2024090\allowbreak3100022028, KQTD20210811090044003), and Scientific Development Fund from Guangdong Provincial Key Laboratory of Visual Media and Multidimensional Intelligence.
\end{acks}

\bibliographystyle{ACM-Reference-Format}
\bibliography{reference}

\end{document}